\documentclass[letterpaper, 10 pt, conference]{ieeeconf}  

\IEEEoverridecommandlockouts                              

\usepackage{graphics} 
\usepackage{epsfig} 
\usepackage{times} 

\usepackage{cite}
\usepackage{booktabs}
\usepackage{amsmath,amssymb,amsfonts}
\usepackage{algorithmic}
\usepackage{setspace}
\usepackage{dblfloatfix} 
\usepackage{subcaption}
\usepackage{graphicx}
\usepackage{textcomp}
\usepackage[table]{xcolor}
\usepackage{multirow}
\usepackage{threeparttable}
\usepackage{tabularx}
\usepackage{mathtools}
\usepackage[free-standing-units=true]{siunitx}
\usepackage{float}
\usepackage{placeins}
\usepackage{pifont}
\usepackage[T1]{fontenc} 

\title{\LARGE \bf
ActiveVLN: Towards Active Exploration via \\ Multi-Turn RL in Vision-and-Language Navigation
}

\author{Zekai Zhang$^{1,*}$, Weiye Zhu$^{1,*}$, Hewei Pan$^{1}$, Xiangchen Wang$^{1}$, Rongtao Xu$^{2,3}$, Xing Sun$^{4}$, Feng Zheng$^{1,2,\dagger}$
\thanks{* Equal contribution.}
\thanks{$\dagger$ Corresponding author.}
\thanks{$^{1}$Southern University of Science and Technology, China.}%
\thanks{$^{2}$Spatialtemporal AI, China.}%
\thanks{$^{3}$MBZUAI, UAE.}
\thanks{$^{4}$Tencent Youtu Lab.}
}

\begin{document}

\maketitle
\thispagestyle{empty}
\pagestyle{empty}

\begin{abstract}
The Vision-and-Language Navigation (VLN) task requires an agent to follow natural language instructions and navigate through complex environments.  
Existing MLLM-based VLN methods primarily rely on imitation learning (IL) and often use DAgger for post-training to mitigate covariate shift. While effective, these approaches incur substantial data collection and training costs.  
Reinforcement learning (RL) offers a promising alternative. However, prior VLN RL methods lack dynamic interaction with the environment and depend on expert trajectories for reward shaping, rather than engaging in open-ended active exploration. This restricts the agent’s ability to discover diverse and plausible navigation routes.  
To address these limitations, we propose ActiveVLN, a VLN framework that explicitly enables active exploration through multi-turn RL. In the first stage, a small fraction of expert trajectories is used for IL to bootstrap the agent. In the second stage, the agent iteratively predicts and executes actions, automatically collects diverse trajectories, and optimizes multiple rollouts via the GRPO objective.  
To further improve RL efficiency, we introduce a dynamic early-stopping strategy to prune long-tail or likely failed trajectories, along with additional engineering optimizations.  
Experiments show that ActiveVLN achieves the largest performance gains over IL baselines compared to both DAgger-based and prior RL-based post-training methods, while reaching competitive performance with state-of-the-art approaches despite using a smaller model.
Code and data will be released soon.
 
\end{abstract}

\section{Introduction}

\PARstart{V}{ision}-and-Language Navigation (VLN) \cite{r2r,vlnce} requires a navigation agent to interpret natural language instructions and navigate toward target locations.
With the recent progress of multimodal large language models (MLLMs), VLN has entered a new stage. By leveraging their seamless integration of language understanding and visual perception, MLLM-based approaches have achieved substantial improvements over conventional VLN models. Currently, VLN methods using a monocular RGB camera \cite{navid, uni-navid, navila, streamvln} mostly adopt this paradigm, employing MLLMs as the backbone for action decision making.

Conventional MLLM-based VLN methods primarily adopt imitation learning (IL), where agents mimic expert demonstrations to map observations to actions. However, IL suffers from covariate shift, where errors accumulate when the agent encounters states unseen during training, leading to poor generalization. To mitigate this, many works~\cite{navid, uni-navid, navila, streamvln} employ DAgger \cite{dagger} post-training, which augments training with expert corrections but still requires substantial expert data collection and retraining. Reinforcement learning (RL) offers a natural alternative by enabling agents to learn from self-generated trajectories. Yet, prior RL approaches \cite{vln-r1, octonav} lack dynamic interaction with the environment, relying on expert trajectories for reward shaping instead of engaging in open-ended active exploration, which limits their exploratory capacity and yields only marginal gains over IL.

To address these limitations, we propose \textbf{ActiveVLN}, a VLN framework that explicitly enable active exploration through multi-turn RL. ActiveVLN consists of two training stages. In the first stage, IL is used for bootstrapping, but with far less expert data compared with existing IL-based approaches. In the second stage, the agent performs multi-turn RL, iteratively refining its navigation policy through active exploration without relying on expert trajectories. Specifically, the agent iteratively predicts actions, executes them in the environment, and appends new observations to its input sequence until termination. Multiple rollouts are then optimized with a GRPO objective. To further enhance RL efficiency, we propose a dynamic early-stop strategy that prunes long-tail or likely failed trajectories, along with several engineering optimizations. Beyond the framework, we provide a comprehensive analysis of critical design choices in VLN RL, such as reward shaping and multi-turn paradigm, offering insights into why prior RL approaches often underperform.

We evaluate ActiveVLN on the R2R \cite{r2r} and RxR \cite{rxr} benchmarks in continuous environments. The active exploration process yields substantial gains (+11.6 SR on R2R and +9.7 SR on RxR over IL baselines), enabling ActiveVLN to surpass both DAgger-based and prior RL-based post-training methods, while also achieving competitive performance with state-of-the-art approaches using a smaller model, less training time, and lower data collection costs.

In summary, the main contributions are: 
\begin{itemize}
    \itemsep0em
    \item We propose ActiveVLN, a VLN framework that leverages active exploration to refine navigation policies, allowing agents to learn from diverse, self-generated trajectories with minimal expert supervision.
    \item We develop several optimization techniques, including a dynamic early-stopping strategy, to improve RL training efficiency.
    \item We show that ActiveVLN achieves the largest performance gains over IL baselines compared to both DAgger-based and prior RL-based post-training methods and reaching competitive performance with state-of-the-art approaches despite using a smaller model.
\end{itemize}

    \begin{figure*}[t]
        \centering
        \captionsetup{font=small}
        \includegraphics[width=\linewidth]{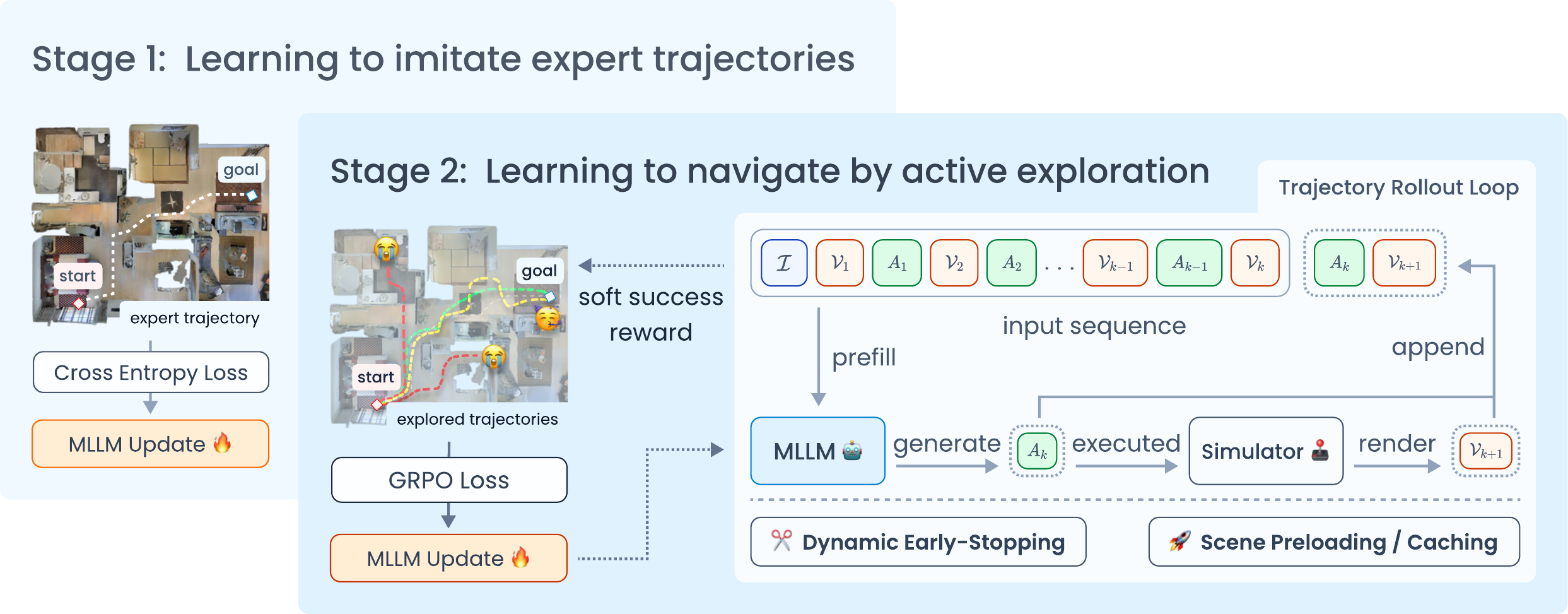}
        \caption{Overview of ActiveVLN. In Stage 1, ActiveVLN performs imitation learning (IL) using expert trajectories. In Stage 2, it conducts multi-turn reinforcement learning (RL), autonomously collecting trajectories in the simulator, receiving rewards that encourage progress toward the goal, and updating the policy via GRPO. Key components, including the \emph{dynamic early-stopping strategy}, \emph{scene preloading}, and \emph{scene caching}, are incorporated to ensure efficient training during RL.}
        \label{fig:overview}
        \vspace{-15pt}
    \end{figure*}

\section{Related Work}
\subsection{Vision-and-Language Navigation}
Vision-and-Language Navigation (VLN) tasks require a navigation agent to navigate through an environment to a target location by following natural language instructions~\cite{r2r}. 
Early VLN studies\cite{scalevln,etpnav,hong2022bridging,vln-petl} primarily focused on discrete environments, where the navigation space is abstracted as a topological graph. In this setting, agents can only move between predefined waypoints on the graph~\cite{r2r, chen2022think, wang2021structured,long2024discuss}. While this abstraction simplifies planning, it significantly diverges from real-world navigation, where both actions and observations are continuous.
To address this gap, Vision-and-Language Navigation in Continuous Environments (VLN-CE) has been proposed~\cite{vlnce, raychaudhuri2021language, wang2023dreamwalker,irshad2021hierarchical}. In VLN-CE, instead of jumping between discrete waypoints, the agent executes low-level actions such as moving forward or rotating. This setting makes it a more realistic and challenging task for embodied navigation.

\subsection{Imitation Learning in VLN}
Early studies~\cite{shah2022lmnav,instructnav, huang2022visual, zhou2024navgpt, vlfm} explored training-free approaches that incorporate large language models (LLMs) or multimodal large language models (MLLMs) into navigation pipelines, leveraging their strong capabilities in language understanding and visual perception. 
However, such modular frameworks often suffer from misalignment between components, and their performance heavily depends on the pre-trained LLMs or MLLMs. 
To overcome these limitations, recent work~\cite{navid, uni-navid, navila} has investigated end-to-end training with LLM~\cite{llama2,llama3,qwen2} and MLLM backbones~\cite{llava_video,nvila}, where imitation learning (IL) has become the dominant paradigm. In this setting, navigation policies are trained on large-scale expert trajectories to map visual observations to low-level actions. 
Yet, naive IL suffers from covariate shift. Since the agent is only exposed to expert trajectories during training, small errors at test time can accumulate when encountering unseen states, leading to compounding deviations from the goal. 
To mitigate this, DAgger~\cite{dagger} has been widely adopted~\cite{navid, uni-navid, navila, streamvln}, where additional expert corrections are collected on visited states to improve generalization and robustness. 
Despite their effectiveness, IL-based methods remain heavily dependent on large-scale expert data, restricting the agent’s ability to explore diverse and plausible routes. They also incur substantial data collection and retraining costs, as each iteration requires new annotations and policy updates. 
In contrast, our approach empowers the navigation agent to explore actively, enabling it to discover plausible navigation behaviors beyond expert demonstrations, without requiring manual trajectory collection or retraining from scratch.

\subsection{Reinforcement Learning in VLN}
Reinforcement learning (RL) has been successful in improving the reasoning of large language models (LLMs), with methods like Group Relative Policy Optimization (GRPO) being particularly effective~\cite{grpo, deepseek-r1}. Unlike other methods such as Proximal Policy Optimization (PPO)~\cite{ppo}, GRPO simplifies the process by using group-wise reward normalization, which eliminates the need for a separate value model and reduces computational costs.
Despite its success in LLMs, the use of RL in Vision-and-Language Navigation (VLN) is still limited. Current approaches often rely on expert trajectories for reward shaping and lack dynamic interaction with the environment~\cite{vln-r1, octonav}, which restricts the agent's ability to explore different paths.
To address these limitations, we use a multi-turn RL approach that treats VLN as a sequential decision process~\cite{deepeyes, ragen}. This method enables agents to learn optimal policies through direct interaction with the simulator, without needing expert trajectories.
\section{Task Formulation}\label{sec:task-formulation}
Each navigation episode is defined by a tuple $(s_0, \mathcal{I}, s^*)$, where $s_0$ is the agent’s initial state (position and orientation), $\mathcal{I}$ is a natural language instruction to follow, and $s^*$ is the target location.
At each time step $t$, the agent receives a forward-facing monocular RGB image $\mathbf{v}_t$ as its observation and maintains a history $\mathbf{h}_t = \{\mathbf{v}_{1:t}, a_{1:t-1}\}$ of past observations and actions.
A parameterized policy $\pi_\theta$ selects a low-level action $a_t \in \mathcal{A}$ based on the instruction and current history:
\begin{equation}
    a_t \sim \pi_\theta(a_t \mid \mathcal{I}, \mathbf{h}_t).
\end{equation}

The action space is defined as:
\begin{equation}
    \mathcal{A} = \{\, \texttt{FORWARD},\; \texttt{LEFT},\; \texttt{RIGHT},\; \texttt{STOP} \,\},
\end{equation}
where \texttt{FORWARD} moves the agent forward by a fixed step size, and \texttt{LEFT}/\texttt{RIGHT} rotate the agent by a fixed angle, both determined by the task configuration.
An episode terminates when the agent issues \texttt{STOP}. Success is defined as stopping within 3 meters of the goal $s^*$.
\section{Method}
In this section, we first present our formulation of VLN as a multi-turn process (Section~\ref{sec:multi-turn}).  
We then describe the ActiveVLN framework (Section~\ref{sec:ActiveVLN}), which enables learning from self-generated trajectories through active exploration (Section~\ref{sec:exploration}) and employs a dynamic early-stopping strategy to enhance RL training efficiency (Section~\ref{sec:early-stop}).  
Finally, we provide engineering details for further acceleration of RL training (Section~\ref{sec:engineering}).

\subsection{Multi-Turn Paradigm}\label{sec:multi-turn}

\begin{figure}[htbp]
    \centering
    \captionsetup{font=small}
    \begin{subfigure}[b]{\linewidth}
        \centering
        \includegraphics[width=\linewidth]{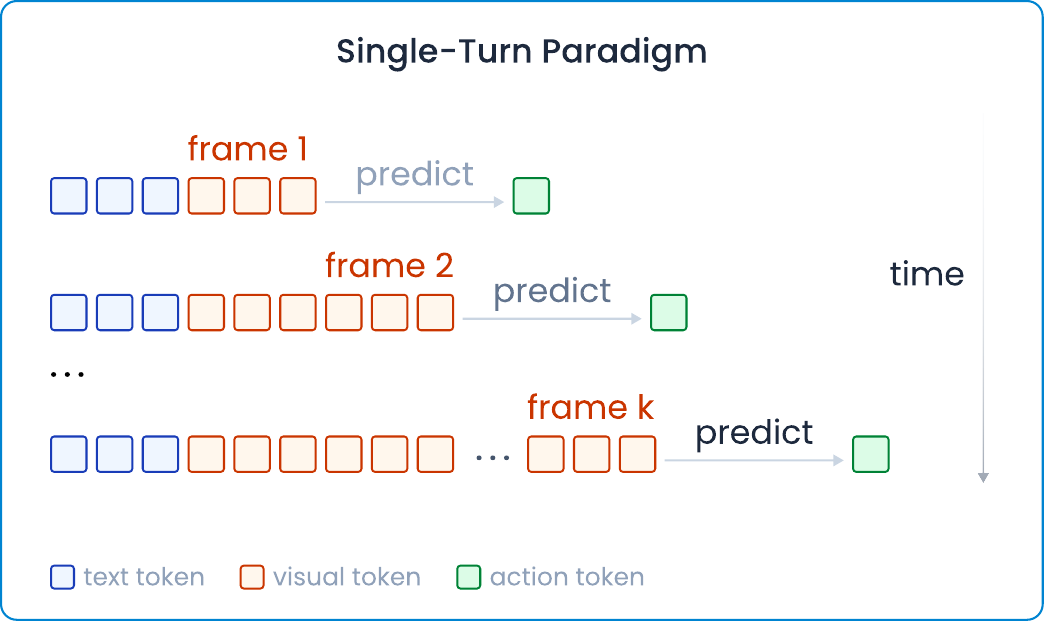}
        \caption{Single-turn paradigm: each action is predicted from the instruction and past observations only.}
        \label{fig:bundled}
    \end{subfigure} \\
    \begin{subfigure}[b]{\linewidth}
        \centering
        \includegraphics[width=\linewidth]{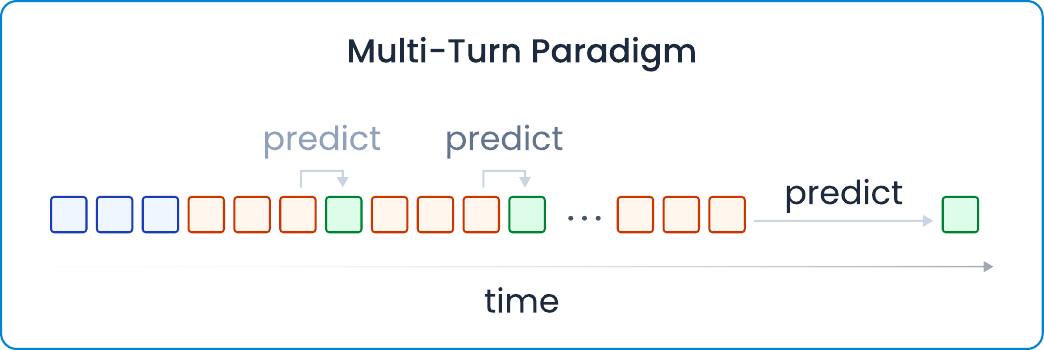}
        \caption{Multi-turn paradigm: actions are generated autoregressively from the instruction, past observations, and past actions. This allows training signals from future steps to backpropagate and refine earlier predictions.}
        \label{fig:streaming}
    \end{subfigure}
    \caption{Comparison between the single-turn and multi-turn paradigms.}
    \label{fig:images}
\end{figure}

Following video-based MLLMs, most prior end-to-end VLN models adopt the \emph{single-turn paradigm} (Figure~\ref{fig:bundled}), where each action is predicted from the instruction and past observations:
\begin{equation}
    a_t \sim \pi_\theta(a_t \mid \mathcal{I}, \mathcal{V}_{1:t}).
\end{equation}

In contrast, we adopt the \emph{multi-turn paradigm} (Figure~\ref{fig:streaming}), where actions are modeled autoregressively from both past observations and actions:
\begin{equation}
    a_t \sim \pi_\theta\left(a_t \mid \mathcal{I}, \{(\mathcal{V}_i, a_i)\}_{i=1}^{t-1}, \mathcal{V}_t\right).
\end{equation}

This paradigm offers several advantages. 
First, it naturally enables KV-cache reuse for efficient inference. 
Second, while the single-turn paradigm breaks actions within the same episode into independent pieces, the multi-turn formulation allows actions to be packed together, accelerating training. 
Most importantly, it enables gradients associated with trajectory outcomes to be propagated to all preceding actions. 
We find this property to be crucial for the success of RL in VLN (see Section~\ref{sec:abl-paradigm}).

\subsection{ActiveVLN Framework}\label{sec:ActiveVLN}

\subsubsection{Compact Action Prediction}
The raw VLN-CE action space comprises four primitive actions: \texttt{FORWARD}, \texttt{TURN LEFT}, \texttt{TURN RIGHT}, and \texttt{STOP}. Prior work augments this space by merging consecutive actions of the same type~\cite{navid,uni-navid,navila}. For instance, three \texttt{FORWARD} steps (each 25cm) can be merged into a single \texttt{FORWARD 75cm} action. This augmentation diversifies action granularity and shortens episode length, improving training efficiency.
Building on this, we adopt action chunking to further reduce episode length, where the agent predicts up to three future actions at once rather than a single action per step.

\subsubsection{Active Exploration}\label{sec:exploration}  
Since MLLMs are not pre-trained for VLN tasks, we start with imitation learning (IL) using a small number of expert demonstrations to initialize the navigation policy.  
The IL objective is:  
\begin{equation}
\mathcal{L}_{\text{IL}} = 
- \sum_{t} \sum_{i=1}^{n_t} \log P(a_t^i \mid \mathcal{I}, \mathcal{H}_{<t}, a_t^{<i}),
\end{equation}
where $\mathcal{I}$ is the navigation instruction,  
$\mathcal{H}_{<t}$ is the interaction history up to step $t$, and $a_t^{<i}$ are the actions already generated in the current chunk.  
Formally, the history is:  
\begin{equation}
\mathcal{H}_{<t} = \{\mathcal{V}_1, A_1, \mathcal{V}_2, A_2, \dots, \mathcal{V}_{t-1}, A_{t-1}, \mathcal{V}_t\},
\end{equation}
where $\mathcal{V}_t$ is the observation at step $t$, and $A_t = \{a_t^1, a_t^2, \dots, a_t^{n_t}\}$ is a sequence of low-level actions.  

IL provides a good starting point, but it has a key limitation: the agent only learns to mimic expert trajectories.  
Once it encounters unfamiliar situations, it has no mechanism to recover or adapt.  
To overcome this, we introduce active exploration through reinforcement learning (RL).  
Here, the agent is no longer restricted to expert data.  
Instead, it can interact with the environment on its own: given an instruction, it predicts an action, executes it in the simulator, and observes the outcome.  
By repeating this loop until it issues a \texttt{stop} action (or an exception occurs), the agent actively generates diverse trajectories, learns from its successes and failures, and gradually improves its policy.  

For optimization, we use Group Relative Policy Optimization (GRPO)~\cite{grpo}.  
GRPO samples $G$ candidate trajectories for the same instruction and compares them within the group.  
Trajectories that perform better than the group average are reinforced, while weaker ones are suppressed.  
The RL objective is:  
\begin{equation}
\begin{aligned}
\mathcal{L}_{\text{RL}} = \mathbb{E}_{\{o_i\}} \left[ \frac{1}{G} \sum_{i=1}^G \frac{1}{|o_i|} \sum_{t=1}^{|o_i|}  
\min \left( \frac{\pi_\theta}{\pi_{\theta_{\text{old}}}} A_{i,t}, \right. \right. \\
\left. \left. \text{clip}\!\left(\frac{\pi_\theta}{\pi_{\theta_{\text{old}}}}, 1-\epsilon, 1+\epsilon\right) A_{i,t} \right) \right],
\end{aligned}
\end{equation}
where $o_i$ is the $i$-th trajectory with length $|o_i|$, $\pi_\theta$ and $\pi_{\theta_{\text{old}}}$ are the new and old policies, $\epsilon$ is the clipping parameter, and $A_{i,t}$ is the estimated advantage.  

We use a \textit{soft success reward}:  
\begin{equation}
R = 15 \cdot \mathbb{I}(\text{success}) \cdot \frac{d_{\text{goal}}}{3},
\end{equation}
where $d_{\text{goal}}$ is the geodesic distance to the goal.  
The indicator $\mathbb{I}(\text{success}) = 1$ if the agent issues a valid \texttt{stop} within 3 meters of the goal, and $0$ otherwise.  
In this way, the agent is no longer just imitating experts but is encouraged to explore actively, discover different ways of reaching the goal, and improve by trial and error.  
This self-driven learning process is the key to stronger generalization in unseen environments.

\subsubsection{Dynamic Early-Stopping Strategy}\label{sec:early-stop}
Trajectory rollout time can account for over half of total RL training time, making it the primary bottleneck. 
We observe that excessively long trajectories often dominate this time, and in most cases correspond to unsuccessful attempts in which the agent wanders aimlessly or explores irrelevant regions. 

To address this issue, we introduce a \emph{dynamic early-stopping strategy} that adaptively terminates unpromising rollouts. Specifically, a trajectory is stopped and marked as failed once its length exceeds a threshold \( T_{\text{max}} \), defined as:
\begin{equation}
    T_{\text{max}} = \alpha_{\text{roll}} \cdot |\tau^*|,
\end{equation}
where \( |\tau^*| \) is the length of the oracle (expert trajectory), and \( \alpha_{\text{roll}} > 1 \) is a tolerance factor that specifies how much deviation from the oracle is acceptable (we set \( \alpha_{\text{roll}} = 2 \) in our experiments).  

This strategy avoids wasting computation on hopeless rollouts, while maintaining a balance between preventing excessive exploration and avoiding overly strict cutoffs, ultimately leading to more efficient training.

\subsection{Engineering Details in RL} \label{sec:engineering}
To further accelerate RL training, we adopt several techniques: \textit{Scene caching} stores frequently accessed scene data in memory, enabling faster loading when the same scene is revisited. \textit{Scene preloading} pipelines scene loading with policy updates, reducing idle time during training. These techniques cut down scene-loading overhead and improve training efficiency. In addition, similar to \cite{ragen}, we decouple the simulator from the training server by deploying it as a standalone HTTP server, which allows scalable and parallel execution of multiple navigation environments.

\section{Experiments}

\subsection{Simulation Benchmark and Evaluation Metrics}
We evaluate the R2R and RxR val-unseen splits on Matterport3D\cite{Matterport3D} scenes in continuous environments using the Habitat simulator\cite{habitat}. Since these scenes are unseen during training, the task requires generalization to novel environments. Following prior work, we report several VLN evaluation metrics, including success rate (SR), oracle success rate (OSR), success weighted by path length (SPL), navigation error (NE) and and normalize dynamic time wrapping
(nDTW)\cite{ndtw}. An episode is considered successful if the agent issues the STOP action within 3m of the goal.

\subsection{Real-World Robot Settings}
We also validate our method in real world using a wheeled humanoid robot. 
RGB observations are captured using the Intel RealSense D435i camera mounted on the robot’s head, with the height adjusted to approximately 1.25\,m to match the simulation setting. 
The trained navigation policy is served remotely and the robot continuously streams observations to the server and executes the control commands returned in response.
 
\subsection{Implementation Details}
\subsubsection{Training Data}
We utilize a total of 167.6k expert trajectories to bootstrap the policy with IL, including 10.8k from R2R\cite{r2r}, 146.3k from R2R-EnvDrop\cite{envdrop}, and 10.5k from RxR\cite{rxr}. To ensure data quality, we filter out RxR samples with excessively long instructions or paths.  
For the subsequent RL post-training, we sample 4k navigation instructions from R2R and RxR, respectively. At this stage, ground-truth actions are not required.	Instead, the agent actively engages in the exploration of diverse navigation routes.

\subsubsection{Hyper-Parameters}
We use Qwen2.5-VL-3B-Instruct \cite{qwen2.5vl} as the base model with the AdamW optimizer \cite{adamw}.  
During the IL stage, only the connector and language model are updated while the vision encoder remains frozen. The peak learning rate is $2\times10^{-5}$ with a global batch size of 64.  
In the RL post-training stage, all parameters are trainable. Each batch includes 8 prompts, each of which would generate 4 trajectory rollouts. The KL coefficient is set to 0.0, the peak learning rate to $1\times10^{-6}$, and the clipping parameter to 0.28. We employ dynamic sampling \cite{dapo} with at most 2 attempts to ensure at least one successful rollout per group. Each action chunk contains up to three merged actions.
All experiments are conducted on 4 NVIDIA L40S GPUs, requiring about 20 hours for IL and RL training, respectively.  

\subsection{Comparisons with SOTA Methods}
\begin{table}[htbp]
    \centering
    \footnotesize
    \setlength{\tabcolsep}{3pt}
        \resizebox{\linewidth}{!}{
        \begin{tabular}{l l r r r r r}
    \toprule
    
    \multirow{2}{*}{Method (Model Size)} & \multirow{2}{*}{Stage} & \multicolumn{5}{c}{R2R Val-Unseen} \\
    \cmidrule(lr){3-7}
    & & NE$\downarrow$ & OS$\uparrow$ & SR$\uparrow$ & SPL$\uparrow$ & $\Delta$SR \\
    \midrule
    \multicolumn{2}{l}{\textbf{DAgger-based post-training}} \\
    \multirow{2}{*}{\hspace{2mm}NaVid (7B) \cite{navid}} & IL & 5.90 & 46.7 & 33.1 & 30.7\\
                                   & IL + DAgger & 5.50 & 49.1 & 37.4 & 35.9 & +4.3\\
    \addlinespace
    \multirow{2}{*}{\hspace{2mm}UniNaVid (7B) \cite{uni-navid}} & IL & - & - & - & -\\
                                   & IL + DAgger & 5.58 & 53.3 & 47.0 & 42.7 & - \\                    
    \addlinespace
    \multirow{2}{*}{\hspace{2mm}MapNav$^\dagger$ (7B) \cite{mapnav}} & IL & 6.38 & 38.2 & 23.9 & 19.5 \\
                     & IL + DAgger  & 6.02 & 46.1 & 33.5  & 30.7 & +9.6 \\
    \addlinespace
    \multirow{2}{*}{\hspace{2mm}StreamVLN$^\dagger$ (7B)  \cite{streamvln}} & IL & 6.05 & 53.8 & 45.5 & 41.6 & \\
                         & IL + DAgger & 5.47 & 57.8 & \textbf{50.8} & \textbf{45.7} & +5.3 \\
    \midrule
    \multicolumn{2}{l}{\textbf{RL-based post-training}} \\
    \multirow{2}{*}{\hspace{2mm}VLN-R1 (7B) \cite{vln-r1}} & IL & 7.92 & 37.1 & 24.9 & 17.5 \\
                            & IL + RL & 7.00 & 41.2 & 30.2 & 21.8 & +5.3 \\
    \addlinespace
    \multirow{3}{*}{\hspace{2mm}\textbf{ActiveVLN (3B, Ours)}} & IL & 6.83 & 42.7 & 38.5 & 33.7 &  \\
                        & IL + DAgger & 6.11 & 50.2 & 45.4 & 41.5 & +6.9 \\
                        & IL + RL & \textbf{5.31} & \textbf{58.3} & 50.1 & 43.7 & \textbf{+11.6} \\
    \bottomrule
    \end{tabular}}
    \captionsetup{font=small}
    \caption{Comparison of state-of-the-art methods on the R2R Val-Unseen split before and after post-training (DAgger or RL). $\Delta$SR indicates the improvement in success rate (SR) after post-training. $^\dagger$ denotes results obtained using only VLN-specific data to ensure a fair comparison.}
    \label{tab:r2r_post_training}
\end{table}

\begin{table}[htbp]
    \centering
    \footnotesize
    \setlength{\tabcolsep}{4pt}
    \begin{tabular}{l c c c c c c}
    \toprule
    \multirow{2}{*}{Method (Model Size)} & \multicolumn{2}{c}{Training data} & 
    \multicolumn{4}{c}{RxR Val-Unseen} \\
    
    \cmidrule(lr){2-3} \cmidrule(lr){4-7}
    & VLN & Others & NE$\downarrow$ & SR$\uparrow$ & SPL$\uparrow$ & nDTW$\uparrow$ \\ 
    \midrule
    NaVILA (8B) \cite{navila}           & -     & \ding{51} & 6.77 & 49.3 & 44.0 & 58.8 \\
    UniNaVid (7B) \cite{uni-navid}   & 2.4M  & \ding{51} & 6.24 & 48.7 & 40.9 & - \\
    StreamVLN (7B) \cite{streamvln}     & 1M    & \ding{51} & 6.22 & \textbf{52.9} & \textbf{46.0} & \textbf{61.9} \\
    \midrule
    ActiveVLN (3B, w/o RL) & 177K  & \ding{55}         & 7.25 & 41.0 & 34.1 & - \\
    \textbf{ActiveVLN (3B, Ours)} & 177K  & \ding{55}         & \textbf{5.84} & 50.7 & 41.2 & 58.1 \\
    \bottomrule
    \end{tabular}
    \captionsetup{font=small}
    \caption{Comparison of state-of-the-art methods on the RxR Val-Unseen split. The columns under “Training data” indicate the amount of VLN data used and whether additional datasets (e.g. visual question-answering) were included.}
    \label{tab:rxr_results}
\end{table}

Table~\ref{tab:r2r_post_training} compares our method with SOTA approaches on the R2R Val-Unseen split, before and after post-training (DAgger or RL). For fairness, results are reported under the same dataset setting (using only VLN-specific data) whenever available in the original papers.
DAgger consistently improves SR across models, but its effect diminishes once the IL baseline is already strong. In contrast, our active exploration delivers the largest performance gain while maintaining SOTA performance, despite requiring less training time and a smaller model. Specifically, SR improves from 38.5 (IL) to 50.1 (IL+RL), yielding a remarkable +11.6 gain that surpasses larger models.  
We also apply DAgger to our IL baseline, and it yields smaller SR gains than our active exploratory RL approach, while requiring additional expert trajectories and longer training time (26h vs. 20h).
These results underscore the effectiveness and efficiency of active exploratory RL post-training, enabling policy refinement without reliance on expert corrections.  

Table~\ref{tab:rxr_results} reports results on the RxR benchmark. Unlike prior methods that are trained not only on VLN data but also on additional data sources such as visual question answering~\cite{zhang2025llavavideo,azuma2022scanqa} and human touring data~\cite{lin2023learning,navila}, ActiveVLN is trained solely on 177K VLN expert trajectories without any extra data. Despite this constraint, ActiveVLN achieves the lowest navigation error (NE=5.84) and a competitive success rate (SR=50.7) compared to StreamVLN, which relies on substantially more data. This highlights that our method achieves strong generalization while remaining data-efficient. Notably, after incorporating active exploration (w/ RL), the success rate (SR) improves by 9.7.
The slightly lower SPL can be attributed to the nature of active exploration, which prioritizes autonomous trajectory discovery over strict imitation of expert demonstrations.

\subsection{Qualitative Results}
\subsubsection{Results in Simulation}
\begin{figure*}[htbp]
    \centering
    \captionsetup{font=small}
    \includegraphics[width=\linewidth]{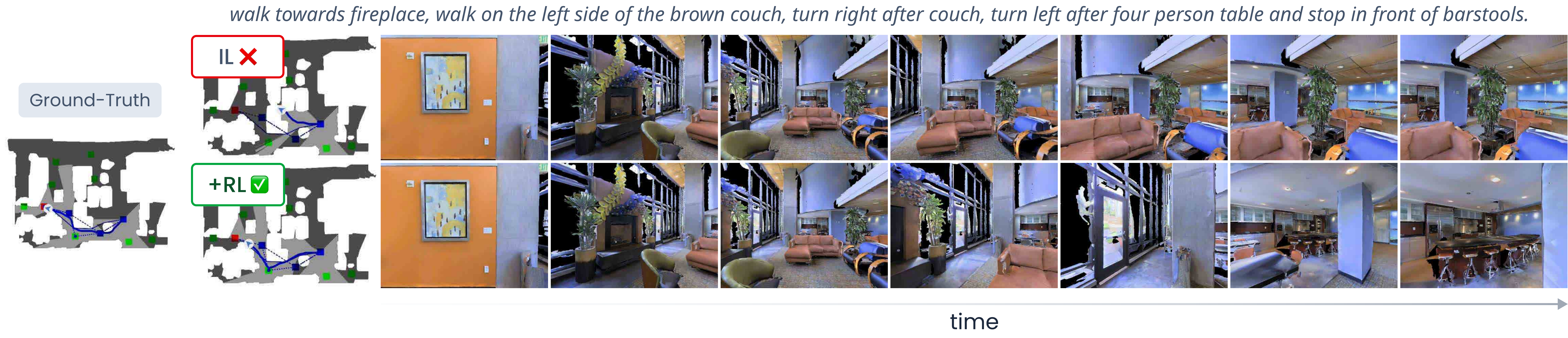}
    \caption{Qualitative results of ActiveVLN in simulation. We compare ActiveVLN with its ablated variant without RL post-training. The IL-only baseline fails to execute the instructed left turn around the brown couch and eventually gets stuck. In contrast, ActiveVLN with RL post-training successfully follows the given instructions, passes on the correct side of the couch, and reaches the goal near the barstools.}
    \label{fig:traj-rgb} 
\end{figure*}

Figure~\ref{fig:traj-rgb} illustrates a representative example where the IL-only baseline fails to reach the goal, whereas the policy with RL post-training succeeds. Specifically, the IL policy does not execute the instructed left turn to navigate around the brown couch. In contrast, the RL-enhanced policy follows the correct path, passes the couch on the appropriate side, and reaches the target successfully.

\subsubsection{Results in Real-World Scenarios}
\begin{figure*}[thbp]
    \centering
    \captionsetup{font=small}
    \includegraphics[width=\linewidth]{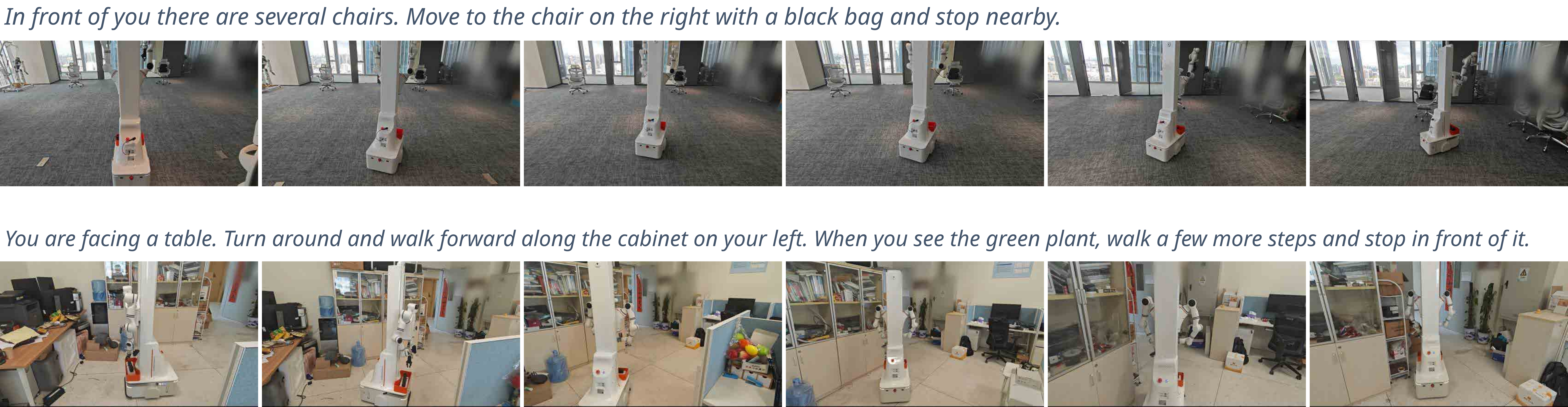}
    \caption{Qualitative results of ActiveVLN in real-world scenarios, including a laboratory environment and an office workspace. Additional examples are provided in the supplementary video.}
    \label{fig:world-result} 
\end{figure*}

We deploy ActiveVLN in diverse real-world scenes, including offices, hallway and laboratories. Representative qualitative results are shown in Figure~\ref{fig:world-result}, where ActiveVLN successfully completes navigation tasks. Full demonstrations are available in the supplementary video.

\subsection{Ablations}
We analyze key design choices in RL post-training, including reward design (Section~\ref{sec:abl-reward}) and the single-turn vs. multi-turn paradigm (Section~\ref{sec:abl-paradigm}). We then assess the effect of dynamic early-stopping (Section~\ref{sec:abl-early-stop}) and other optimization techniques (Section~\ref{sec:abl-others}) on training efficiency.

\subsubsection{Reward Design} \label{sec:abl-reward}
We explore alternative reward designs beyond the \textit{soft success reward}. Note that we do not incorporate dynamic sampling to accelerate training.

The \textit{hard success reward} is defined as:
\begin{equation}
R_{\text{hard}} = 15 \cdot \mathbb{I}(\text{success}),
\end{equation}
which eliminates the distance-to-goal normalization used in the soft success reward.

The \textit{path alignment reward} is formulated as:
\begin{align}
R_{\text{align}} &= w_{\text{succ}} \cdot \mathbb{I}(\text{success})\cdot \frac{d_{\text{to-goal}}}{3} + w_{\text{nDTW}} \cdot \text{nDTW}, \\
\text{where} \quad & w_{\text{succ}} + w_{\text{nDTW}} = 15 \nonumber.
\end{align}
This introduces an additional nDTW term that captures the similarity between the predicted and ground-truth trajectories.

\begin{table}[htbp]
    \centering
    \resizebox{\linewidth}{!}{
    \begin{tabular}{lcccc}
        \toprule
        \multirow{2}{*}{Reward Type} & 
    \multicolumn{4}{c}{R2R Val-Unseen} \\
    \cmidrule(lr){2-5}
         & NE$\downarrow$ & OS$\uparrow$ & SR$\uparrow$ & SPL$\uparrow$ \\
        \midrule
        soft success reward & 5.56 & \textbf{56.4} & \textbf{48.1} & \textbf{40.6} \\ 
        hard success reward & 5.68 & 53.6 & 46.2 & 39.9 \\
        path align. reward (\(w_{\text{succ}}:w_{\text{nDTW}}=\mathbf{3{:}1} \)) & \textbf{5.49} & 51.9 & 46.3 & 40.1 \\
        path align. reward (\(w_{\text{succ}}:w_{\text{nDTW}}=\mathbf{2{:}1} \)) & 5.64 & 48.6 & 44.4 & 40.0 \\
        path align. reward (\(w_{\text{succ}}:w_{\text{nDTW}}=\mathbf{1{:}1} \)) & 5.61 & 49.6 & 44.4 & 40.3 \\
        \bottomrule
    \end{tabular}}
    \captionsetup{font=small}
    \caption{Performance under different reward designs.}
    \label{tab:reward-ablation}
\end{table}
As shown in Table~\ref{tab:reward-ablation}, the soft success reward yields the best overall performance. In contrast, the hard success reward, which removes distance-to-goal normalization, leads to a reduced SR, likely because the normalization encourages the agent to continue moving closer to the goal when it is nearby.
Incorporating the nDTW term via the path alignment reward also leads to performance degradation. Interestingly, the path alignment reward with \( w_{\text{succ}}:w_{\text{nDTW}} = \mathbf{3{:}1} \) achieves the lowest navigation error (NE), suggesting that the nDTW term helps the agent better mimic expert trajectories. However, this comes at the cost of reduced SR, indicating that overemphasizing trajectory similarity can hinder goal completion. This trend is further supported by the fact that increasing the weight on nDTW (from 3:1 to 1:1) leads to a consistent decline in both OS and SR.
These results suggest that the soft success reward, while simple, is an effective training signal that encourages both goal reaching and diverse and effective pathfinding.

\begin{table}[htbp]
    \centering
    \resizebox{\linewidth}{!}{
    \begin{tabular}{ccllll}
    \toprule
    \multicolumn{2}{c}{Stage} & \multicolumn{4}{c}{R2R Val-Unseen} \\
    \cmidrule(lr){1-2}   \cmidrule(lr){3-6}
    IL  &  RL  & NE$\downarrow$ & OS$\uparrow$ & SR$\uparrow$ & SPL$\uparrow$ \\
    \midrule
    \multicolumn{4}{l}{\textbf{Multi-turn Paradigm}} \\
      \hspace{4mm}$\checkmark$   &               &  6.83 & 42.7 & 38.5 & 33.7        \\
      \hspace{4mm}$\checkmark$   & $\checkmark$  &  \textbf{5.31} {\color{gray}(-1.52)} & \textbf{58.3} {\color{gray}(+15.6)} & \textbf{50.1} {\color{gray}(+11.6)} & \textbf{43.7} {\color{gray}(+10.0)} \\      
    \midrule
    \multicolumn{4}{l}{\textbf{Single-turn Paradigm}} \\
      \hspace{4mm}$\checkmark$   &               &  6.44 & 48.1 & 39.7 & 35.3        \\ 
      \hspace{4mm}$\checkmark$   & $\checkmark$  &  6.25 {\color{gray}(-0.19)} & 48.1 {\color{gray}(+0.0)} & 40.3 {\color{gray}(+0.6)} & 37.0 {\color{gray}(+1.7)} \\
    
    \bottomrule
    \end{tabular}}
    \captionsetup{font=small}
    \caption{Performance under the multi-turn and single-turn paradigms. RL post-training provides substantial improvements across all metrics in the multi-turn paradigm, while yielding only marginal gains in the single-turn paradigm.}
    \label{tab:inference_paradigm_ablation}
\end{table}

\subsubsection{Effect of the Multi-Turn Paradigm} \label{sec:abl-paradigm}
To investigate the effect of the multi-turn paradigm, we construct a single-turn baseline and apply RL under the same training data and key hyper-parameter settings as used in the multi-turn version of ActiveVLN. The results are summarized in Table~\ref{tab:inference_paradigm_ablation}.
After RL, the single-turn baseline achieves a slightly higher SR than the multi-turn variant. A plausible explanation is that the single-turn paradigm processes inputs as short video segments, which aligns more closely with the base model’s pretraining on large-scale video datasets. In contrast, the multi-turn paradigm involves longer-horizon, interleaved image–text sequences, which are less common in the pretrained distribution and may initially limit performance.  
However, the multi-turn paradigm demonstrates substantially larger performance improvements after RL. This indicates that RL particularly benefits from the multi-turn formulation, as it enables policy optimization over action histories. While single-turn may initially appear advantageous, multi-turn ultimately provides a stronger foundation for learning complex navigation strategies under RL.

\subsubsection{Effect of Dynamic Early-Stopping Strategy}\label{sec:abl-early-stop}
\begin{figure}[htbp]
    \centering
    \captionsetup{font=small}
    \includegraphics[width=0.85\linewidth]{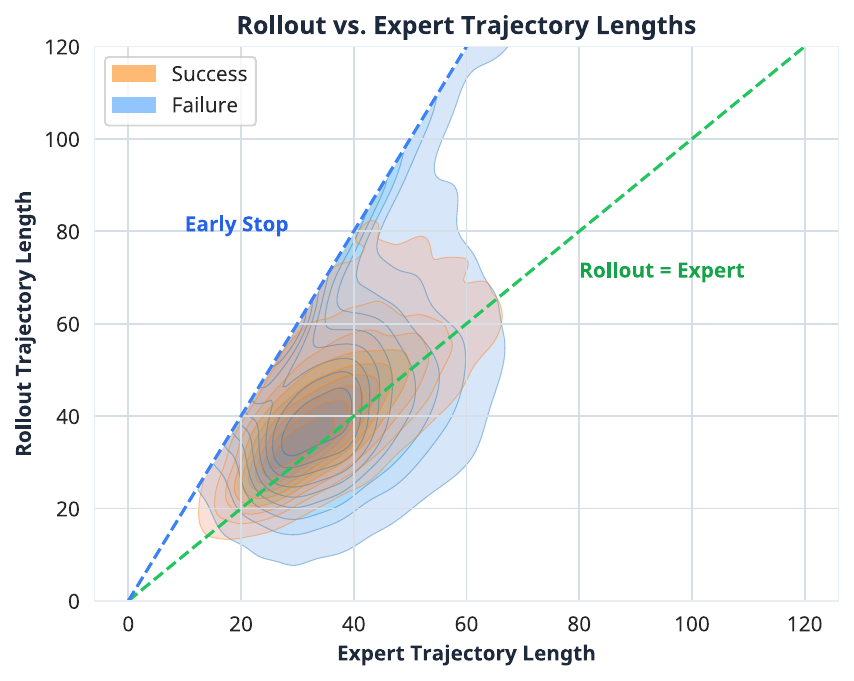}
    \caption{Comparison of rollout and expert trajectory lengths.}
    \label{fig:episode-len-dist} 
\end{figure}

\begin{figure}[thbp]
    \centering
    \captionsetup{font=small}
    \begin{subfigure}[htbp]{0.22\textwidth}
        \includegraphics[width=\linewidth]{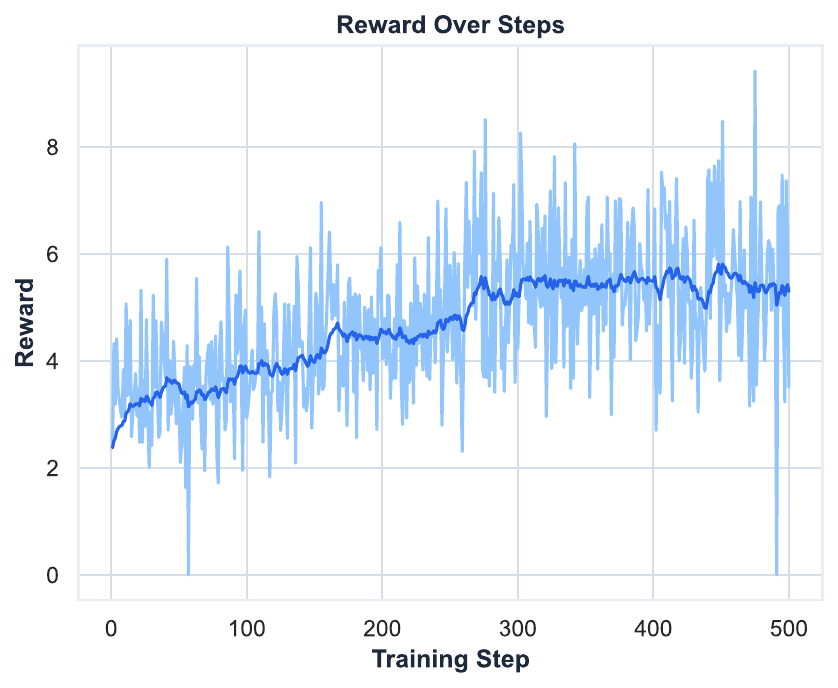}
        \caption{Training Reward}
        \label{fig:reward}
    \end{subfigure}
    \begin{subfigure}[htbp]{0.22\textwidth}
        \includegraphics[width=\linewidth]{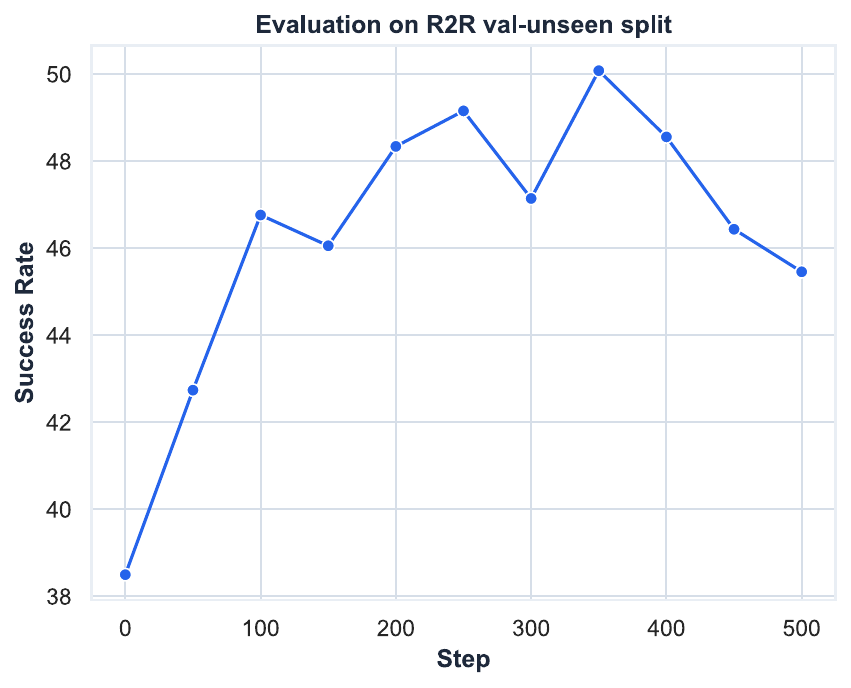}
        \caption{SR on R2R val-unseen split}
        \label{fig:sr}
    \end{subfigure}
    \begin{subfigure}[htbp]{0.22\textwidth}
        \includegraphics[width=\linewidth]{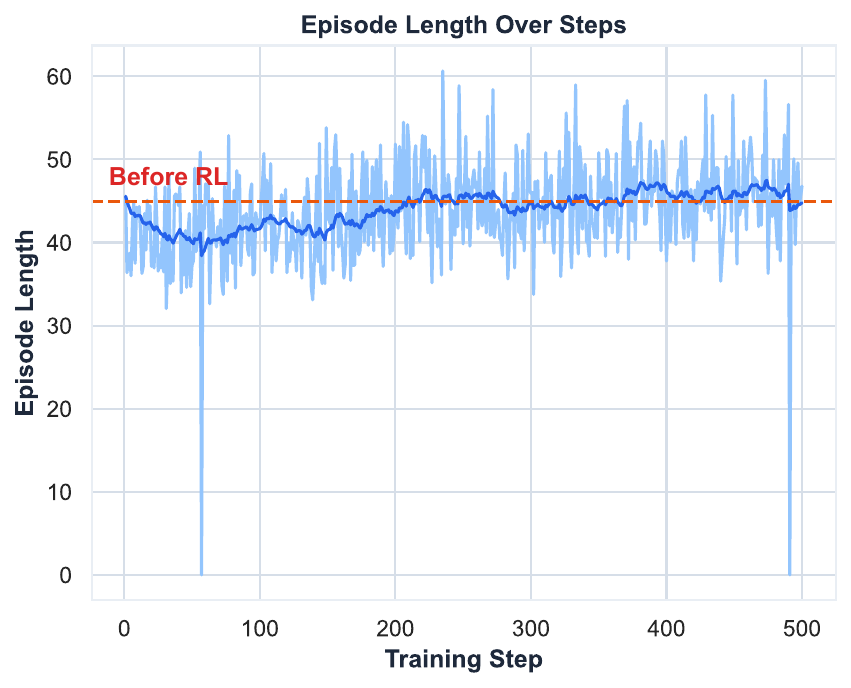}
        \caption{Average Episode Length}
        \label{fig:episode-len}
    \end{subfigure}
    \begin{subfigure}[htbp]{0.22\textwidth}
        \includegraphics[width=\linewidth]{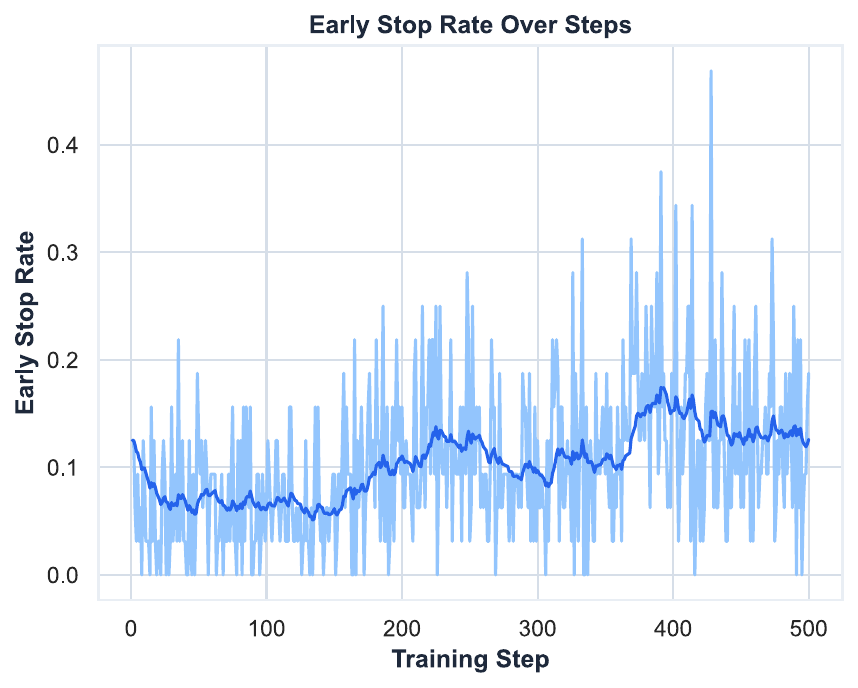}
        \caption{Early Stop Rate}
        \label{fig:early-stop}
    \end{subfigure}
       \caption{Training dynamic of ActiveVLN.}

    \label{fig:training-dynamics}
\end{figure}

Figure~\ref{fig:early-stop} shows that about 10\% of trajectory rollouts are forcefully terminated due to excessive length, reducing the average rollout time by about 10\% (Table~\ref{tab:training_time}).  
Figure~\ref{fig:episode-len-dist} compares the episode length distributions of rollout and expert trajectories. As expected, rollout distributions concentrate around the green reference line (expert trajectories) and become sparser with increasing deviation. The blue line indicates the threshold set by our early-stopping strategy.  
These results demonstrate that the dynamic early-stopping strategy effectively prunes long-tail and failed trajectories while retaining most successful ones, improving both training efficiency and stability.

\subsubsection{Effect of Other Optimization Techniques}\label{sec:abl-others}
\begin{table}[htbp]
    \centering
    \footnotesize
    \setlength{\tabcolsep}{6pt}
    \begin{tabular}{ll}
        \toprule
        Optimization Technique & Avg. Rollout Time (s) \\
        \midrule
        ActiveVLN & \textbf{106.9} \\
        \quad w/o Dynamic Early-Stopping & 118.6 \, (+10.9\%) \\
        \quad w/o Scene Pre-Loading      & 111.9 \, (+4.68\%) \\
        \quad w/o Scene Caching          & 109.3 \, (+2.24\%) \\
        \bottomrule
    \end{tabular}
    \captionsetup{font=small}
    \caption{Ablation study of optimization techniques on rollout time.}
    \label{tab:training_time}
\end{table}
To evaluate the effect of other optimization techniques on RL training efficiency, we measure the average rollout time per step over 25 training steps. The results are reported in Table~\ref{tab:training_time}, where each technique is ablated in turn. 
It shows that removing any of them slows down training, with the dynamic early-stopping strategy providing the largest reduction in rollout time. These underscore the importance of our proposed optimizations in significantly improving RL training efficiency.

\subsection{Training Dynamics}

\begin{figure}[htbp]
    \centering
    \captionsetup{font=small}
    \includegraphics[width=\linewidth]{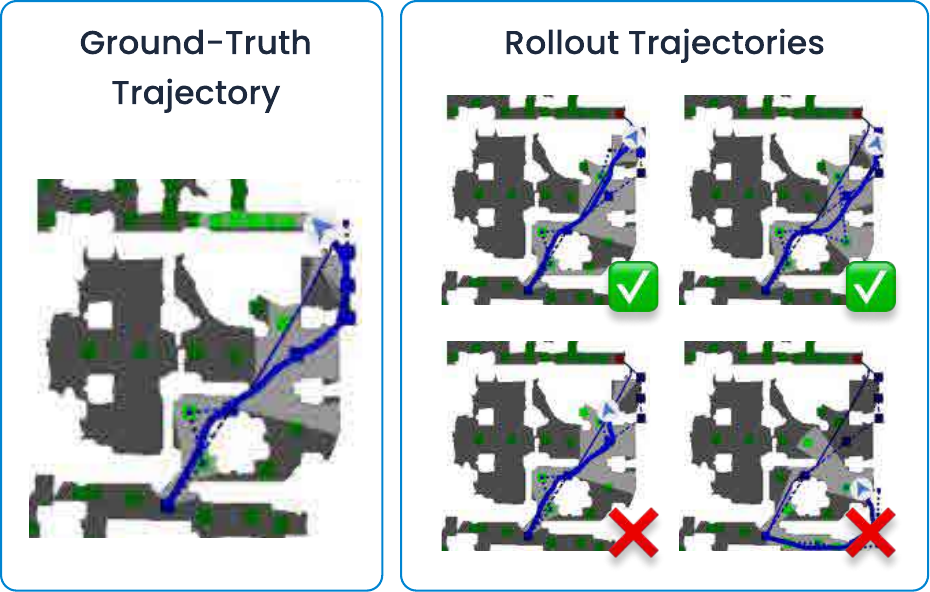}
    \caption{Top-down views of ground-truth and rollout trajectories. The top two cases show successful navigation via alternative paths deviating from the expert trajectory. The bottom two illustrate failures: stopping too far from the goal (bottom left) and moving in the wrong direction from the start (bottom right).}
    \label{fig:rollout-traj} 
\end{figure}

To gain insights into policy behavior during RL post-training, we monitor several key metrics throughout training, as shown in Figure~\ref{fig:training-dynamics}.  
Figure~\ref{fig:reward} shows that reward steadily increases, converging around step 300, reflecting improved performance. This trend aligns with the success rate in Figure~\ref{fig:sr}, evaluated on R2R Val-Unseen every 50 steps, which peaks around step 350 before slightly declining.  
Figure~\ref{fig:episode-len} reports episode length (number of actions per trajectory). After an initial decrease due to format errors causing premature terminations (steps 1-50), episode length gradually approaches the IL baseline as the policy learns correct action formatting.  
Additionally, Figure~\ref{fig:rollout-traj} presents top-down visualizations of a group of rollout trajectories during RL post-training, qualitatively illustrating the benefits of incorporating active exploration: the agent learns to follow diverse paths that deviate from expert demonstrations while still successfully reaching the goal.

\section{Conclusion}
In this work, we present ActiveVLN, a VLN framework that explicitly enables active exploration through multi-turn RL.  
This active exploration process allows agents to learn from diverse, self-generated trajectories without heavy reliance on expert supervision.  
We further introduce a dynamic early-stopping strategy and several engineering optimizations to improve RL efficiency.  
Experiments show that ActiveVLN achieves the largest performance gains over IL baselines compared to both DAgger-based and prior RL-based post-training methods and reaching competitive performance with state-
of-the-art approaches despite using a smaller model. 
These results highlight the critical role of active exploration in VLN and validate our RL post-training framework.
Future work may explore more advanced exploration strategies and extend ActiveVLN to other embodied tasks.


\FloatBarrier
\bibliographystyle{bibliography/IEEEtran}
\bibliography{bibliography/references}

\begin{thebibliography}{10}
\providecommand{\url}[1]{#1}
\csname url@rmstyle\endcsname
\providecommand{\newblock}{\relax}
\providecommand{\bibinfo}[2]{#2}
\providecommand\BIBentrySTDinterwordspacing{\spaceskip=0pt\relax}
\providecommand\BIBentryALTinterwordstretchfactor{4}
\providecommand\BIBentryALTinterwordspacing{\spaceskip=\fontdimen2\font plus
\BIBentryALTinterwordstretchfactor\fontdimen3\font minus \fontdimen4\font\relax}
\providecommand\BIBforeignlanguage[2]{{%
\expandafter\ifx\csname l@#1\endcsname\relax
\typeout{** WARNING: IEEEtran.bst: No hyphenation pattern has been}%
\typeout{** loaded for the language `#1'. Using the pattern for}%
\typeout{** the default language instead.}%
\else
\language=\csname l@#1\endcsname
\fi
#2}}

\bibitem{r2r}
P.~Anderson, Q.~Wu, D.~Teney, J.~Bruce, M.~Johnson, N.~S{\"u}nderhauf, I.~Reid, S.~Gould, and A.~Van Den~Hengel, ``Vision-and-language navigation: Interpreting visually-grounded navigation instructions in real environments,'' in \emph{CVPR}, 2018.

\bibitem{vlnce}
J.~Krantz, E.~Wijmans, A.~Majundar, D.~Batra, and S.~Lee, ``Beyond the nav-graph: Vision and language navigation in continuous environments,'' in \emph{ECCV}, 2020.

\bibitem{navid}
J.~Zhang, K.~Wang, R.~Xu, G.~Zhou, Y.~Hong, X.~Fang, Q.~Wu, Z.~Zhang, and H.~Wang, ``Navid: Video-based vlm plans the next step for vision-and-language navigation,'' in \emph{RSS}, 2024.

\bibitem{uni-navid}
J.~Zhang, K.~Wang, S.~Wang, M.~Li, H.~Liu, S.~Wei, Z.~Wang, Z.~Zhang, and H.~Wang, ``Uni-navid: A video-based vision-language-action model for unifying embodied navigation tasks,'' in \emph{RSS}, 2025.

\bibitem{navila}
A.-C. Cheng, Y.~Ji, Z.~Yang, X.~Zou, J.~Kautz, E.~Biyik, H.~Yin, S.~Liu, and X.~Wang, ``Navila: Legged robot vision-language-action model for navigation,'' in \emph{RSS}, 2025.

\bibitem{streamvln}
M.~Wei, C.~Wan, X.~Yu, T.~Wang, Y.~Yang, X.~Mao, C.~Zhu, W.~Cai, H.~Wang, Y.~Chen, \emph{et~al.}, ``Streamvln: Streaming vision-and-language navigation via slowfast context modeling,'' \emph{arXiv preprint arXiv:2507.05240}, 2025.

\bibitem{dagger}
S.~Ross, G.~Gordon, and D.~Bagnell, ``A reduction of imitation learning and structured prediction to no-regret online learning,'' in \emph{AISTATS}, 2011.

\bibitem{vln-r1}
Z.~Qi, Z.~Zhang, Y.~Yu, J.~Wang, and H.~Zhao, ``Vln-r1: Vision-language navigation via reinforcement fine-tuning,'' \emph{arXiv preprint arXiv:2506.17221}, 2025.

\bibitem{octonav}
C.~Gao, L.~Jin, X.~Peng, J.~Zhang, Y.~Deng, A.~Li, H.~Wang, and S.~Liu, ``Octonav: Towards generalist embodied navigation,'' \emph{arXiv preprint arXiv:2506.09839}, 2025.

\bibitem{rxr}
A.~Ku, P.~Anderson, R.~Patel, E.~Ie, and J.~Baldridge, ``Room-across-room: Multilingual vision-and-language navigation with dense spatiotemporal grounding,'' in \emph{EMNLP}, 2020.

\bibitem{scalevln}
Z.~Wang, J.~Li, Y.~Hong, Y.~Wang, Q.~Wu, M.~Bansal, S.~Gould, H.~Tan, and Y.~Qiao, ``Scaling data generation in vision-and-language navigation,'' in \emph{CVPR}, 2023.

\bibitem{etpnav}
D.~An, H.~Wang, W.~Wang, Z.~Wang, Y.~Huang, K.~He, and L.~Wang, ``Etpnav: Evolving topological planning for vision-language navigation in continuous environments,'' \emph{TPAMI}, 2024.

\bibitem{hong2022bridging}
Y.~Hong, Z.~Wang, Q.~Wu, and S.~Gould, ``Bridging the gap between learning in discrete and continuous environments for vision-and-language navigation,'' in \emph{CVPR}, 2022.

\bibitem{vln-petl}
Y.~Qiao, Z.~Yu, and Q.~Wu, ``Vln-petl: parameter-efficient transfer learning for vision-and-language navigation,'' in \emph{ICCV}, 2023.

\bibitem{chen2022think}
S.~Chen, P.-L. Guhur, M.~Tapaswi, C.~Schmid, and I.~Laptev, ``Think global, act local: Dual-scale graph transformer for vision-and-language navigation,'' in \emph{CVPR}, 2022.

\bibitem{wang2021structured}
H.~Wang, W.~Wang, W.~Liang, C.~Xiong, and J.~Shen, ``Structured scene memory for vision-language navigation,'' in \emph{CVPR}, 2021.

\bibitem{long2024discuss}
Y.~Long, X.~Li, W.~Cai, and H.~Dong, ``Discuss before moving: Visual language navigation via multi-expert discussions,'' in \emph{ICRA}, 2024.

\bibitem{raychaudhuri2021language}
S.~Raychaudhuri, S.~Wani, S.~Patel, U.~Jain, and A.~X. Chang, ``Language-aligned waypoint (law) supervision for vision-and-language navigation in continuous environments,'' in \emph{EMNLP}, 2021.

\bibitem{wang2023dreamwalker}
H.~Wang, W.~Liang, L.~Van~Gool, and W.~Wang, ``Dreamwalker: Mental planning for continuous vision-language navigation,'' in \emph{CVPR}, 2023.

\bibitem{irshad2021hierarchical}
M.~Z. Irshad, C.-Y. Ma, and Z.~Kira, ``Hierarchical cross-modal agent for robotics vision-and-language navigation,'' in \emph{ICRA}, 2021.

\bibitem{shah2022lmnav}
D.~Shah, B.~Osinski, B.~Ichter, and S.~Levine, ``{LM}-nav: Robotic navigation with large pre-trained models of language, vision, and action,'' in \emph{CoRL}, 2022.

\bibitem{instructnav}
Y.~Long, W.~Cai, H.~Wang, G.~Zhan, and H.~Dong, ``Instructnav: Zero-shot system for generic instruction navigation in unexplored environment,'' in \emph{CoRL}, 2024.

\bibitem{huang2022visual}
C.~Huang, O.~Mees, A.~Zeng, and W.~Burgard, ``Visual language maps for robot navigation,'' in \emph{ICRA}, 2023.

\bibitem{zhou2024navgpt}
G.~Zhou, Y.~Hong, Z.~Wang, X.~E. Wang, and Q.~Wu, ``Navgpt-2: Unleashing navigational reasoning capability for large vision-language models,'' in \emph{ECCV}, 2024.

\bibitem{vlfm}
N.~Yokoyama, S.~Ha, D.~Batra, J.~Wang, and B.~Bucher, ``Vlfm: Vision-language frontier maps for zero-shot semantic navigation,'' in \emph{ICRA}, 2024.

\bibitem{llama2}
H.~Touvron, L.~Martin, K.~Stone, P.~Albert, A.~Almahairi, Y.~Babaei, N.~Bashlykov, S.~Batra, P.~Bhargava, S.~Bhosale, \emph{et~al.}, ``Llama 2: Open foundation and fine-tuned chat models,'' \emph{arXiv preprint arXiv:2307.09288}, 2023.

\bibitem{llama3}
A.~Grattafiori, A.~Dubey, A.~Jauhri, A.~Pandey, A.~Kadian, A.~Al-Dahle, A.~Letman, A.~Mathur, A.~Schelten, A.~Vaughan, \emph{et~al.}, ``The llama 3 herd of models,'' \emph{arXiv preprint arXiv:2407.21783}, 2024.

\bibitem{qwen2}
Q.~Team, ``Qwen2 technical report,'' \emph{arXiv preprint arXiv:2407.10671}, 2024.

\bibitem{llava_video}
Y.~Zhang, J.~Wu, W.~Li, B.~Li, Z.~Ma, Z.~Liu, and C.~Li, ``Video instruction tuning with synthetic data,'' \emph{arXiv preprint arXiv:2410.02713}, 2024.

\bibitem{nvila}
Z.~Liu, L.~Zhu, B.~Shi, Z.~Zhang, Y.~Lou, S.~Yang, H.~Xi, S.~Cao, Y.~Gu, D.~Li, \emph{et~al.}, ``Nvila: Efficient frontier visual language models,'' in \emph{CVPR}, 2025.

\bibitem{grpo}
Z.~Shao, P.~Wang, Q.~Zhu, R.~Xu, J.~Song, X.~Bi, H.~Zhang, M.~Zhang, Y.~Li, Y.~Wu, \emph{et~al.}, ``Deepseekmath: Pushing the limits of mathematical reasoning in open language models,'' \emph{arXiv preprint arXiv:2402.03300}, 2024.

\bibitem{deepseek-r1}
D.~Guo, D.~Yang, H.~Zhang, J.~Song, R.~Zhang, R.~Xu, Q.~Zhu, S.~Ma, P.~Wang, X.~Bi, \emph{et~al.}, ``Deepseek-r1: Incentivizing reasoning capability in llms via reinforcement learning,'' \emph{arXiv preprint arXiv:2501.12948}, 2025.

\bibitem{ppo}
J.~Schulman, F.~Wolski, P.~Dhariwal, A.~Radford, and O.~Klimov, ``Proximal policy optimization algorithms,'' \emph{arXiv preprint arXiv:1707.06347}, 2017.

\bibitem{deepeyes}
Z.~Zheng, M.~Yang, J.~Hong, C.~Zhao, G.~Xu, L.~Yang, C.~Shen, and X.~Yu, ``Deepeyes: Incentivizing" thinking with images" via reinforcement learning,'' \emph{arXiv preprint arXiv:2505.14362}, 2025.

\bibitem{ragen}
Z.~Wang, K.~Wang, Q.~Wang, P.~Zhang, L.~Li, Z.~Yang, X.~Jin, K.~Yu, M.~N. Nguyen, L.~Liu, \emph{et~al.}, ``Ragen: Understanding self-evolution in llm agents via multi-turn reinforcement learning,'' \emph{arXiv preprint arXiv:2504.20073}, 2025.

\bibitem{Matterport3D}
A.~Chang, A.~Dai, T.~Funkhouser, M.~Halber, M.~Niessner, M.~Savva, S.~Song, A.~Zeng, and Y.~Zhang, ``Matterport3d: Learning from rgb-d data in indoor environments,'' in \emph{3DV}, 2017.

\bibitem{habitat}
M.~Savva, A.~Kadian, O.~Maksymets, Y.~Zhao, E.~Wijmans, B.~Jain, J.~Straub, J.~Liu, V.~Koltun, J.~Malik, \emph{et~al.}, ``Habitat: A platform for embodied ai research,'' in \emph{CVPR}, 2019.

\bibitem{ndtw}
G.~Ilharco, V.~Jain, A.~Ku, E.~Ie, and J.~Baldridge, ``General evaluation for instruction conditioned navigation using dynamic time warping,'' in \emph{NeurIPS}, 2019.

\bibitem{envdrop}
H.~Tan, L.~Yu, and M.~Bansal, ``Learning to navigate unseen environments: Back translation with environmental dropout,'' in \emph{NAACL}, 2019.

\bibitem{qwen2.5vl}
S.~Bai, K.~Chen, X.~Liu, J.~Wang, W.~Ge, S.~Song, K.~Dang, P.~Wang, S.~Wang, J.~Tang, \emph{et~al.}, ``Qwen2. 5-vl technical report,'' \emph{arXiv preprint arXiv:2502.13923}, 2025.

\bibitem{adamw}
I.~Loshchilov and F.~Hutter, ``Decoupled weight decay regularization,'' in \emph{ICLR}, 2019.

\bibitem{dapo}
Q.~Yu, Z.~Zhang, R.~Zhu, Y.~Yuan, X.~Zuo, Y.~Yue, W.~Dai, T.~Fan, G.~Liu, L.~Liu, \emph{et~al.}, ``Dapo: An open-source llm reinforcement learning system at scale,'' \emph{arXiv preprint arXiv:2503.14476}, 2025.

\bibitem{mapnav}
L.~Zhang, X.~Hao, Q.~Xu, Q.~Zhang, X.~Zhang, P.~Wang, J.~Zhang, Z.~Wang, S.~Zhang, and R.~M. Xu, ``A novel memory representation via annotated semantic maps for vlm-based vision-and-language navigation,'' \emph{arXiv preprint arXiv:2502.13451}, 2025.

\bibitem{zhang2025llavavideo}
Y.~Zhang, J.~Wu, W.~Li, B.~Li, Z.~MA, Z.~Liu, and C.~Li, ``{LL}a{VA}-video: Video instruction tuning with synthetic data,'' \emph{TMLR}, 2025.

\bibitem{azuma2022scanqa}
D.~Azuma, T.~Miyanishi, S.~Kurita, and M.~Kawanabe, ``Scanqa: 3d question answering for spatial scene understanding,'' in \emph{CVPR}, 2022.

\bibitem{lin2023learning}
K.~Lin, P.~Chen, D.~Huang, T.~H. Li, M.~Tan, and C.~Gan, ``Learning vision-and-language navigation from youtube videos,'' in \emph{ICCV}, 2023.

\end{thebibliography}

\end{document}